\documentclass{article}
\usepackage{ijcai25}

\usepackage{times}
\usepackage{soul}
\usepackage{url}
\usepackage[hidelinks]{hyperref}
\usepackage[utf8]{inputenc}
\usepackage[small]{caption}
\usepackage{graphicx}
\usepackage{amsmath}
\usepackage{amsthm}
\usepackage{subfigure}
\usepackage{booktabs}
\usepackage{algorithm}
\usepackage{algorithmic}
\usepackage[switch]{lineno}

\title{Can Breath Biomarkers Causally Influence Blood Glucose? Investigating VOC-Mediated Modulation in Diabetes}

\author{
Varsha Sharma$^1$
\and
Prasanta K. Guha$^2$\and
Avik Ghose$^1$\\
\affiliations
$^1$TCS Research, India\\
$^2$Department of E\&ECE, IIT Kharagpur, India\\
\emails
sharma.varsha1@tcs.com,
pkguha@ece.iitkgp.ac.in,
avik.ghose@tcs.com
}

\begin{document}
\maketitle

\begin{abstract}
Diabetes is a global health burden, and early detection is critical for timely intervention. This study explores a non-invasive, data-driven framework to identify individuals at risk of diabetes using Volatile Organic Compounds (VOCs) and lifestyle variables. We use causal inference techniques to estimate the impact of VOCs such as acetone, isopropanol, isoprene, and ethanol on blood glucose levels. Additionally, we designed a classifier to distinguish diabetics from non-diabetics using  non-invasive markers. We created a risk-based ranking system for individuals in the ``gray zone," and identified natural clusters in the population using Gaussian Mixture Model. Our results suggest that specific VOCs exhibit a strong causal influence on glucose levels and that machine learning models can reliably classify and stratify individuals at high risk. This integrated causal-explainable analysis can support the development of tool for non-invasive early screening of diabetes.
\end{abstract}

\section{Introduction}
Diabetes mellitus represents a growing global health challenge, with an estimated 537 million adults affected worldwide as of 2021 \cite{sun2022idf}. Early detection and intervention are critical to prevent complications, yet current diagnostic methods such as fasting blood glucose tests and HbA1c measurements are invasive, intermittent, and often fail to capture early metabolic dysregulation \cite{tabak2012prediabetes}, \cite{american20242}.  There is an urgent need for non-invasive, real-time biomarkers that can identify at-risk individuals before clinical onset and provide insights into the underlying physiological mechanisms driving glucose dysregulation \cite{teo2021global}.

Recent advances in metabolomics have highlighted breath volatile organic compounds (VOCs) as promising candidates for metabolic monitoring. Breath VOCs like acetone \cite{wang2009study}, \cite{galassetti2005breath}, isoprene \cite{los2020382}, ethanol \cite{galassetti2005breath}, and isopropanol \cite{li2017exhaled} reflect distinct metabolic pathways (ketosis, cholesterol metabolism, and gut microbiome activity) \cite{wang2009study}, \cite{minh2012clinical}, \cite{roy2024machine} and collectively enhance non-invasive glucose monitoring \cite{righettoni2013correlations}.
While prior studies have established correlations between specific VOCs and diabetes, The causal relationship between these biomarkers and blood glucose modulation has not yet been fully elucidated. Correlation alone does not imply causation, and confounding factors such as medication use, lifestyle, and comorbidities may obscure true biological mechanisms.

Several prior studies have examined the association between lifestyle factors and diabetes risk using machine learning and statistical methods. For example, \cite{witarsyah2025causal} modeled diabetes risk from daily habits and diet, while \cite{molina2021deciphering} focused on Mediterranean diet adherence. \cite{he2019causalbg} highlighted the role of physical activity in glycemic control, and \cite{nemat2023causality} used wearable sensors to capture behavioral patterns for risk prediction. Others, such as \cite{prosperi2020causal} and \cite{noh2025diabetes}, explored causal modeling and interpretable ML with lifestyle and exposome data. However, these approaches largely overlook the causal role of VOCs, focusing instead on behavioral and physiological variables. The potential for VOCs as non-invasive causal indicators of diabetes onset remains underexplored.

This study addresses that gap by applying causal inference techniques to determine whether breath-based VOCs actively influence blood glucose levels. In addition, we combine this with supervised and unsupervised machine learning approaches to identify high-risk individuals and uncover latent subgroups. Our key contributions are:
\begin{enumerate}
\item A causal framework for estimating the effect of VOCs on glucose levels.
\item A classification and risk-ranking system to flag high-risk non-diabetic subjects (latent subgroups).
\item A clustering analysis for segmenting diabetic and non-diabetic subjects.
\end{enumerate}

\section{Methods}
To explore the underlying cause-effect relationships between breath VOCs and blood glucose levels, we employed a structured causal inference framework. The goal was to identify whether individual VOCs or their combination causally influence glucose concentration, supporting their use as non-invasive biomarkers for early diabetes detection.

\subsection{Data Description and Preprocessing}

The study involved 94 participants: 32 healthy, 8 newly diagnosed diabetics, 27 with well-controlled diabetes, and 27 with poorly controlled diabetes. Ethical approval was obtained from the Institutional Ethics Committee of IPGMER Hospital, India. Participant recruitment, clinical evaluation, and sample collection were conducted by medical professionals under expert supervision. Informed consent was obtained from all subjects, who refrained from alcohol and tobacco for at least 7 days and rested 30 minutes prior to sampling.

Key variables such as age, gender, BMI, blood pressure, lifestyle habits (sleep, fruits per day, alcohol and tobacco consumption behavior), stress, comorbidities, and medications  (metformin, etc.) were documented. Blood samples were analyzed for glucose and ketone levels, while breath samples (0.5L) were collected using the isothermal re-breathing method and analyzed via Gas Chromatography-Mass Spectrometry (GC-MS) for VOC profiling. Details of participant characteristics, sampling procedures and preprocessing steps are available in \cite{Shar2501:Feasibility}.

\subsection{Causal Association}
We investigated the following causal hypotheses:
\begin{enumerate}
    \item Does individual VOCs causally influence blood glucose levels?

    \item What is the combined causal effect of VOCs on blood glucose?

    \item Are VOCs responsive to changes in blood glucose (reverse causality)?

\end{enumerate}







\subsubsection{Causal Inference Framework}
The framework begins with preprocessing and encoding of treatment variables, followed by constructing causal graphs and using propensity-based techniques to adjust for confounding. For each VOC (acetone, isopropanol, isoprene, ethanol), we modeled it as the treatment variable and blood glucose as the outcome. We employed the DoWhy model to estimate the average treatment effect (ATE) using backdoor criterion-based adjustment and linear regression as the base estimator. To validate these effects, we performed placebo-based refutation tests by permuting treatment assignments and re-estimating the effect, assessing whether the original causal estimate could be attributed to spurious correlations.

In addition to individual VOCs, we also modeled the joint effect of all four VOCs as a multivariate treatment vector to assess cumulative influence on glucose levels. Furthermore, we applied reverse causality models to test whether glucose could be influencing VOC levels instead essential to rule out bidirectional dependencies and confirm causal directionality. 
All causal models were refuted using placebo interventions and observed to maintain effect significance.

\subsubsection{Explainability and Confounding Checks}
To enhance the interpretability of our causal models and validate the relative importance of features influencing blood glucose, we integrated SHapley Additive exPlanations (SHAP) \cite{lundberg2017unified} into our analytical pipeline. 
Using a tree-based ensemble model (XGBoost) trained on VOCs and lifestyle variables, we computed SHAP values to visualize features that strongly influenced the model’s glucose predictions.

Further, to assess the robustness of our causal estimates against varying confounding conditions, we systematically experimented with multiple sets of potential confounders and re-estimated the causal effects of individual VOCs (acetone, isopropanol, isoprene, and ethanol) as well as their combined influence on blood glucose. 

\subsection{Classification and Risk Ranking}

To evaluate the predictive power of non-invasive biomarkers identified through causal modeling, we trained classification models to distinguish diabetic and non-diabetic subjects using only VOCs and lifestyle data (excluding blood glucose and ketone). Models such as logistic regression, XGBoost, and random forest were employed. Furthermore, we generated a risk continuum by ranking predicted probabilities of diabetes, thereby identifying high-risk individuals in the non-diabetic group, forming a clinically important ``gray zone" for early intervention.

To validate the causal influence of breath VOCs on blood glucose levels and their potential in identifying high-risk non-diabetic individuals (gray zone), we created a synthesized glucose marker (synthetic glucose) as a linear combination of four VOCs (acetone, isopropanol, isoprene, and ethanol). The coefficients for the synthetic glucose were derived directly from the causal effect estimates obtained through our causal inference analysis, thereby embedding physiological relevance into the marker. We then used the Mann–Whitney U test to statistically compare synthetic glucose between ``gray zone" individuals and other non-diabetic individuals.

\subsection{Clustering Analysis}
To identify underlying patterns in the data without relying on direct blood markers such as glucose or ketone, we applied clustering using VOC and lifestyle features alone. The optimal number of clusters was determined using the Bayesian Information Criterion (BIC). A Gaussian Mixture Model (GMM) was found to best capture the distributional structure of the data. To visualize these groupings, we applied Principal Component Analysis (PCA) for dimensionality reduction. Notably, the resulting clusters corresponded well with diabetic and non-diabetic groupings.

\section{Results and Validations}

\subsection{Causal Inference}

\begin{table*}
    \centering
    \begin{tabular}{llllll}
        \hline
        Q & Treatment  & Outcome & ATE  & Refute Effect & p-value \\
        \hline
        1  & Acetone     & Glucose & 5.400  & -5.655 $\times 10^{-12}$ & 0.0\\
        2  & Isopropanol & Glucose & -2.8 & -4.888  $\times 10^{-12}$ & 0.0 \\
        3  & Isoprene    & Glucose & 23.086 &  -1.818 $\times 10^{-11}$ & 0.0\\
        4  & Ethanol     & Glucose & 67.109 & -1.0658 $\times 10^{-11}$ & 0.0 \\
        5  & All 4 VOCs  & Glucose & 69.892 & -2.2168 $\times 10^{-12}$ & 0.0 \\
        \hline
         &     & &Reverse Causality   &  \\
         \hline
        1  & Glucose & Acetone     & 0.0275  & -0.00031 & 0.94\\
        2  & Glucose & Isopropanol & -0.0012 &  -3.934 $\times 10^{-6}$ & 0.9\\
        3  & Glucose & Isoprene    & 5.70 $\times 10^{-6}$ & -4.959 $\times 10^{-6}$ & 0.9 \\
        4  & Glucose & Ethanol     & 0.00039 & -1.9166 $\times 10^{-5}$ & 0.9 \\
        5  & Glucose & All 4 VOCs  & 0.0253 & 0.000607 & 0.94 \\
        \hline
    \end{tabular}
    \caption{Summary of Estimated Causal Effects of VOCs on Blood Glucose with Refutation Tests}
    \label{tab:plain}
\end{table*}

We conducted a series of causal inference experiments to quantify the direct impact of VOCs (acetone, isopropanol, isoprene, and ethanol) on blood glucose levels. The analysis was grounded in a forward causal framework, where each VOC was treated as a potential intervention (treatment), and blood glucose was modeled as the outcome. Table \ref{tab:plain} shows that the estimated average treatment effects (ATEs) revealed that all four VOCs had statistically significant causal relationships with glucose levels. Among them, ethanol exhibited the strongest individual causal effect, with an estimated increase of approximately $67.1 mg/dL$ in glucose concentration. Isoprene also showed a notable positive effect $(23.09 mg/dL)$, while acetone contributed modestly $(5.4 mg/dL)$. Interestingly, isopropanol was the only VOC that demonstrated a slight negative causal effect $(-2.8 mg/dL)$, suggesting a potentially protective or inverse relationship probably because when in higher concentrations it gets metabolized to acetone \cite{li2017exhaled}. When all four VOCs were jointly considered as a treatment group, the combined causal effect on glucose levels rose to $69.89 mg/dL$; exceeding any single compound's influence and pointing toward potential synergistic effects.  This supports our hypothesis that changes in VOCs precede changes in blood glucose and not vice versa.

To ensure the robustness of our causal claims, we applied placebo-based refutation tests. These tests involve randomly shuffling the treatment assignments to break any real causal relationships. For all forward causal models, the refuted effects were close to zero and statistically insignificant, while the original causal effects remained highly significant (p-value = $0.0$). This validates that the observed relationships are unlikely to be due to chance or unmeasured confounding. Furthermore, we conducted reverse causality tests by switching the roles of treatment and outcome, modeling glucose as the treatment and VOCs as outcomes. These models yielded near-zero estimated effects and p-values of $0.9$, strongly suggesting the absence of reverse causal influence. 

\subsubsection{Explainability and Confounding Checks}
\begin{figure}[htbp]
\centerline{\includegraphics[width=1\columnwidth]{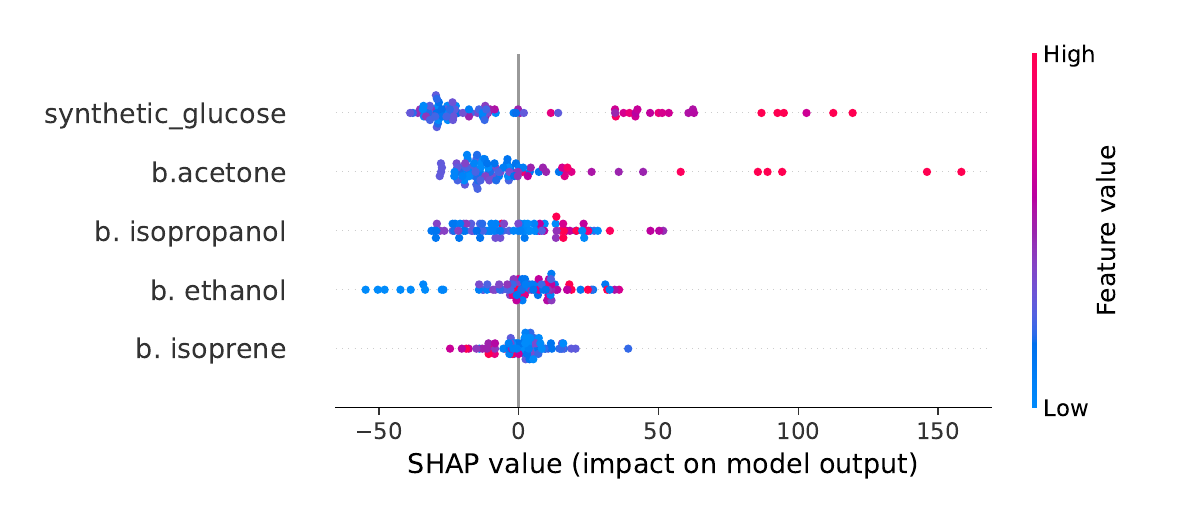}}
\caption{SHAP Summary Plot for VOCs }
\label{fig:shap}
\end{figure}

The SHAP summary plot in Figure \ref{fig:shap} highlights the importance of the causally-informed synthetic glucose feature, which integrates the weighted contributions of four VOCs. Positioned at the top of the plot, synthetic glucose shows a wide SHAP value range, with higher values (red) strongly increasing the model’s predicted diabetes risk, confirming its dominant role. Among the individual VOCs, ethanol and isoprene contribute positively at higher concentrations, while isopropanol exhibits a mostly negative influence, aligning with its inverse causal relationship to glucose. These patterns validate the causal effect estimates and demonstrate that both the composite and individual VOCs meaningfully impact non-invasive diabetes risk prediction.

Further, we define confounder sensitivity as the percentage change in estimated causal effect when varying the included confounders. Using different combinations of confounders such as age, gender, height, weight, co-morbidities, alcohol and tobacco consumption behavior, fruit intake, sleep duration, and stress. We observed minimal sensitivity for acetone (1\%) and isopropanol (5\%), while isoprene and ethanol showed higher variability (20\%). The combined causal effect of all four VOCs changed by 8\%, indicating overall stability. These results demonstrate the robustness of our model and highlight the importance of systematically evaluating confounder influence in causal inference workflows.

\subsection{Classification and Risk Ranking}
\begin{figure}[htbp]
\centerline{\includegraphics[width=0.7\columnwidth]{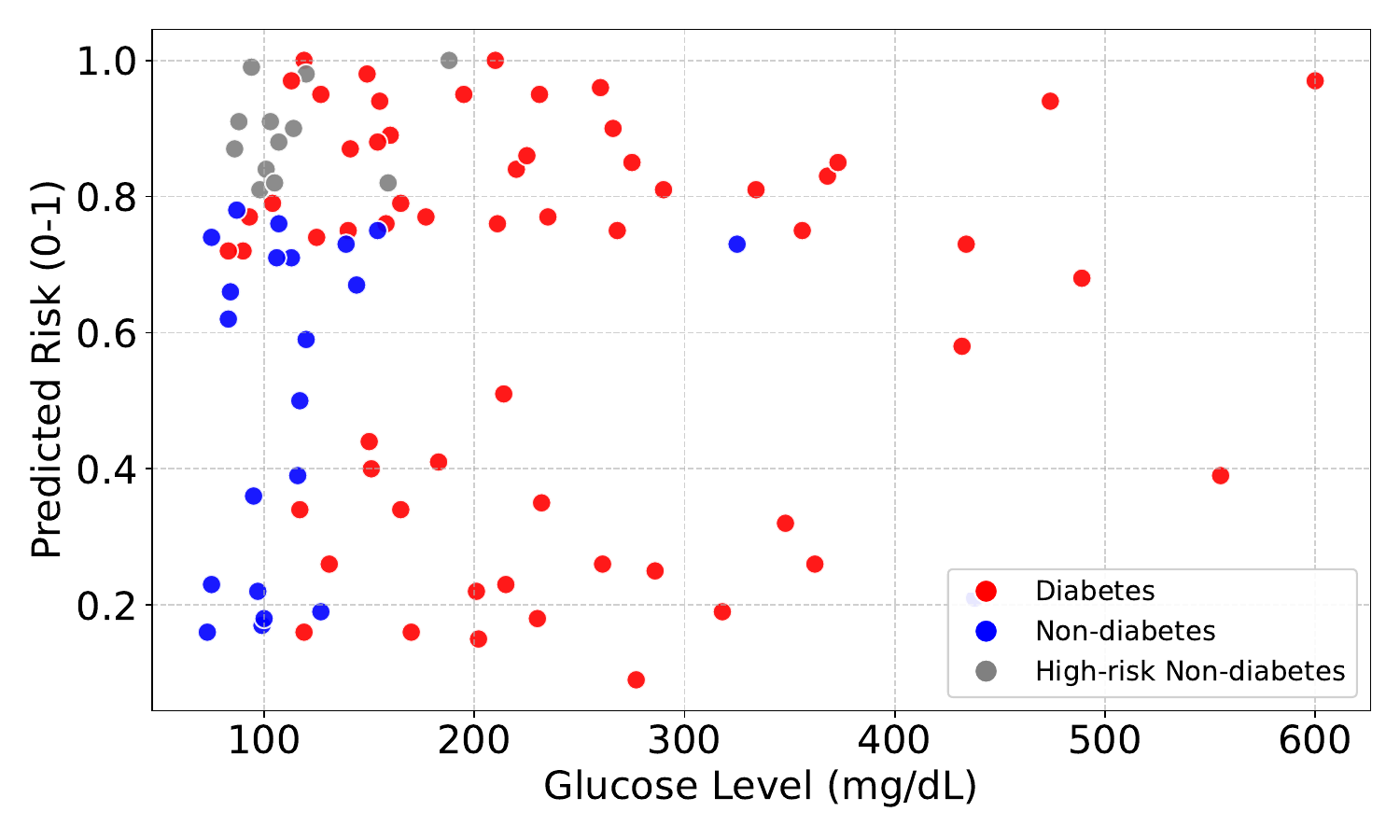}}
\caption{Glucose Levels vs Predicted Risk}
\label{fig:Glu_risk}
\end{figure}

A machine learning classifier was trained using 5-fold cross-validation to predict diabetes status (binary: yes/no) based solely on VOC profiles, excluding traditional blood biomarkers.
\begin{table}[h]
\centering
\scalebox{1}{
\begin{tabular}{@{}llllll@{}}
\toprule
\textbf{Method} &  AUC &  precision &  recall & f1-score \\ \midrule

Classification & 0.938 & 0.942 & 0.936 & 0.934 \\
Clustering & 0.88 & 0.890 &  0.882 & 0.883 \\
\bottomrule
\end{tabular}}
\caption{Performance in Separating Diabetic and Non-Diabetic Subjects}
\label{tab:result_comparison}
\end{table}
Model performance (refer Table \ref{tab:result_comparison}) was evaluated via F1 scores, achieving 94\% accuracy with outliers and 98\% after outlier removal, demonstrating VOC signatures' predictive power. Additionally, a risk continuum model ranked subjects by diabetes probability to identify a ``gray zone" of high-risk non-diabetics (shown in Figure \ref{fig:Glu_risk}), validated by a known prediabetic individual ranked third. Further, validation by Mann–Whitney U test yielded a U statistic of 73.0 with a p-value of 0.035, indicating a statistically significant elevation in VOC-based glucose levels among the high-risk group. This supports the notion that the gray zone group exhibits VOC patterns consistent with increased diabetes risk, captured via non-invasive breath biomarkers.

In contrast, when we conducted the same test using actual measured blood glucose levels, the Mann–Whitney U test yielded a U statistic of 135.0 and a non-significant p-value of 0.93. This suggests that while blood glucose values alone do not distinguish the ``gray zone" from the rest of the non-diabetic population, the VOC-based synthetic glucose, informed by causal analysis, is more sensitive in capturing early metabolic dysregulation. These results highlight the added value of causal modeling in identifying latent physiological risk signatures not readily observable through conventional markers, reinforcing the potential of breath VOCs as early non-invasive indicators for diabetes risk stratification.

\subsection{Clustering Analysis}

\begin{figure}[htbp]
\centerline{\includegraphics[width=0.69\columnwidth]{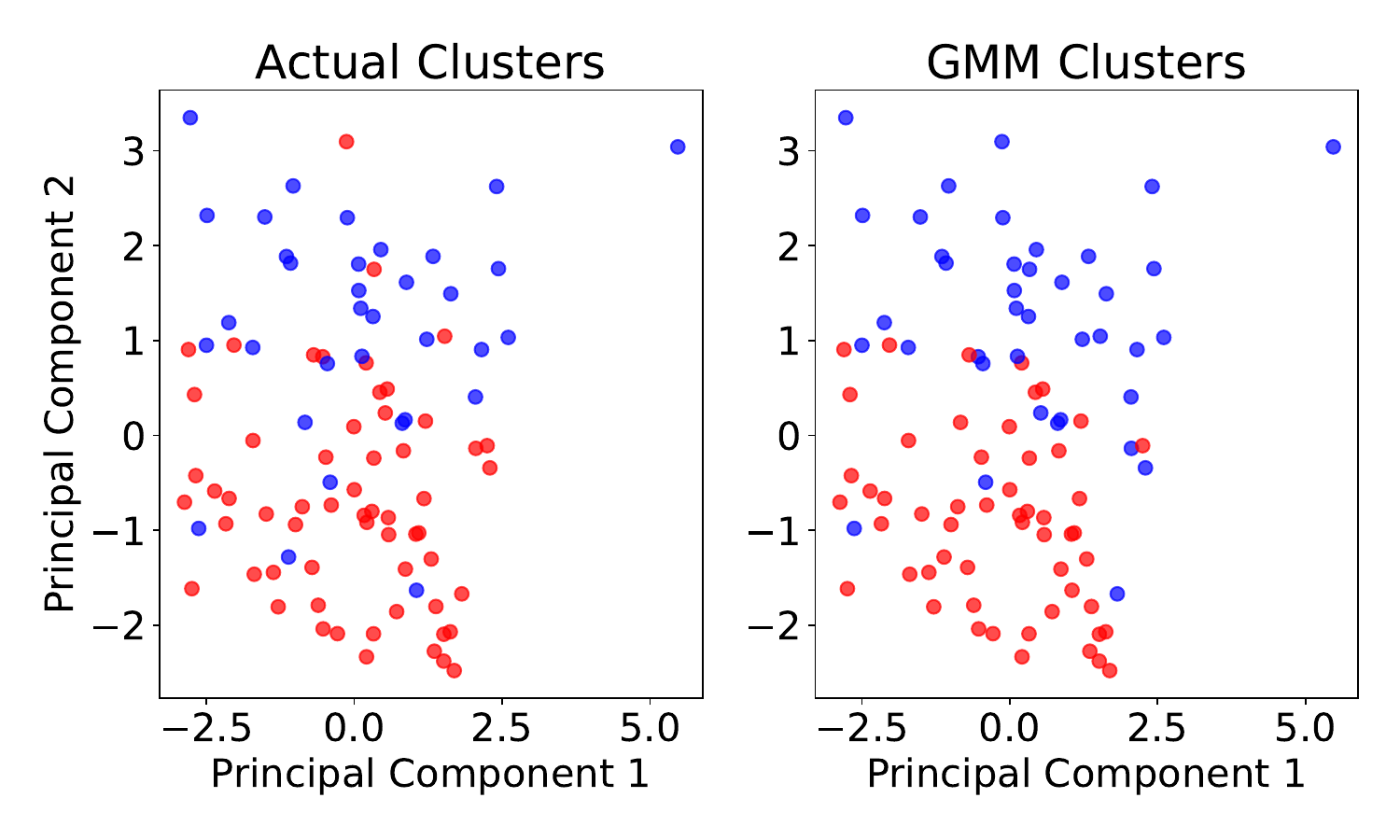}}
\caption{Actual vs GMM Cluster [Red:Diabetes; Blue:Non-Diabetes]}
\label{fig:pairwise_scatter}
\end{figure}

The clustering analysis identified two optimal clusters based on BIC and AIC, aligning well with diabetic and non-diabetic groups as show in Figure \ref{fig:pairwise_scatter}. Clustering quality metrics included a silhouette score of 0.12 (moderate separation), Adjusted Rand Index (ARI) of 0.58, and Normalized Mutual Information (NMI) of 0.48, indicating good agreement with ground truth. ARI reflects the similarity between predicted and true labels, while NMI measures shared information, higher values indicate better clustering. The unsupervised clusters also achieved an 88\% F1 score in distinguishing diabetic status using only VOC and lifestyle features, supporting their utility for non-invasive risk stratification as shown in Table \ref{tab:result_comparison}.

\section{Discussion and Implications}
Our findings indicate that specific VOCs, particularly acetone and isopropanol, exhibit a measurable causal effect on blood glucose levels, reinforcing their potential role as non-invasive biomarkers for assessing diabetes risk. The ability to detect such causal relationships using observational data strengthens the argument for incorporating breath-based VOC sensing into early diagnostic frameworks for metabolic disorders. Importantly, the use of causal inference methods beyond conventional correlational analysis allows for stronger, counterfactual reasoning about the effects of these biomarkers on physiological outcomes. The integration of classification and clustering methods further complements our causal pipeline by enabling patient stratification and risk profiling.

From a physiological standpoint, the observed effects are aligned with established metabolic knowledge \cite{saasa2018sensing}. For example, diabetic individuals often exhibit elevated breath acetone due to increased fat metabolism and ketone production under insulin-deficient conditions, such as in diabetic ketoacidosis (DKA). This biochemical linkage provides a plausible mechanistic explanation for the statistical associations and further supports the clinical validity of VOCs as proxies for metabolic imbalance.

\section{ Conclusion and Future Scope}

In summary, our study presents a unified, explainable causal inference framework that not only supports non-invasive diabetes screening but also offers transparency and rigor suitable for clinical integration. The convergence of machine learning, causal inference, and explainability tools positions our approach as a promising step toward next-generation digital diagnostics.

While our current analysis establishes a causal link between certain VOCs (acetone, ethanol, isoprene, and isopropanol) and elevated blood glucose levels, it remains essential to recognize the presence of underlying clinical confounders, such as insulin deficiency. Both elevated glucose and VOC emission may be manifestations of the same latent physiological process (e.g., ketosis), which is currently unobserved in purely data-driven causal models. Therefore, future work should focus on designing knowledge-driven causal models that integrate clinical domain knowledge to uncover such latent pathophysiological variables. Moreover, validation on larger and more demographically diverse populations is critical to assess the robustness and generalizability of these findings. Another promising direction is the development of ontology-based models that trace the biochemical origins of VOCs, ensuring that the presence of compounds like acetone in breath is accurately attributed to glucose-related metabolic states, such as ketosis. Such advancements would significantly strengthen the clinical relevance and deployment potential of VOC-based, non-invasive diagnostic systems.

\section*{Acknowledgments}
I gratefully acknowledge Souradeep Roy from IIT Kharagpur for his valuable support in data collection, which has significantly contributed to the advancement of this work.

\bibliographystyle{named}
\bibliography{ijcai25}

\end{document}